\documentclass{article}

\usepackage[final,nonatbib]{tackling_climate_workshop_style}

\usepackage[utf8]{inputenc} 
\usepackage[T1]{fontenc}    
\usepackage[colorlinks]{hyperref}       
\usepackage{url}            
\usepackage{booktabs}       
\usepackage{amsfonts}       
\usepackage{nicefrac}       
\usepackage{microtype}      
\usepackage{amsmath}
\usepackage{amssymb}
\usepackage{wrapfig}
\usepackage{floatrow}
\usepackage{graphicx}
\usepackage{float}
\usepackage{xcolor}
\usepackage[ruled,vlined]{algorithm2e}
\usepackage{algorithmic}
\usepackage[size=small,labelfont=bf]{caption}
\renewcommand{\S}{\mathbf{S}}
\newcommand{\s}{\mathbf{s}}
\newcommand{\X}{\mathbf{X}}

\renewcommand{\P}{\mathbb{P}}
\newcommand{\series}[1]{\{#1\}_{t\ge0}}
\newcommand{\R}{\mathbb{R}}

\usepackage[normalem]{ulem}

\title{Identifying Distributional Differences in Convective Evolution Prior to Rapid Intensification in Tropical Cyclones}

\author{%
  Trey McNeely$^{\, 1}$\ \ Galen Vincent$^{\, 1}$\ \ Rafael Izbicki$^{\, 2}$\ \ Kimberly M. Wood$^{\, 3}$\ \ Ann B. Lee$^{\, 1}$\\
  \\
  $^{1}$Department of Statistics \& Data Science, Carnegie Mellon University \\
  $^{2}$Department of Statistics, Federal University of S\~ao Carlos, S\~ao Carlos, Brazil \\
  $^{3}$Department of Geosciences, Mississippi State University \\
  \texttt{imcneely@stat.cmu.edu} \\
}

\begin{document}

\maketitle
\begin{abstract}
Tropical cyclone (TC) intensity forecasts are issued by human forecasters who evaluate spatio-temporal observations (e.g., satellite imagery) and model output (e.g., numerical weather prediction, statistical models) to produce  forecasts every 6 hours. Within these time constraints, it can be challenging to draw insight from such data. While high-capacity machine learning methods are well suited for prediction problems with complex sequence data, extracting interpretable scientific information with such methods is difficult. Here we leverage powerful AI prediction algorithms and classical statistical inference to identify patterns in the evolution of TC convective structure leading up to the rapid intensification of a storm, hence providing forecasters and scientists with key insight into TC behavior.
\end{abstract}

\section{Introduction}
\textbf{Background.} Tropical cyclones (TCs) are powerful, organized atmospheric systems that draw energy from the upper levels of the ocean. They rank among the deadliest and costliest natural disasters in the United States, and rising ocean heat content promises to amplify their danger over time \cite{Klotzbach2018}. TC intensity and motion forecasting alike have improved since the 1990s, but intensity forecasting has lagged behind \cite{DeMaria2014}. Cases of rapid intensification (RI), defined for this work as an intensification of at least 25 knots within 24 hours, are especially difficult to predict \cite{kaplan2003large,Kaplan2010,kaplan2015rii,wood2015definition}. Some of the most impactful TCs in recent years underwent RI prior to landfall (e.g., Hurricanes Irma [2017], Dorian [2019], and Ida [2021]).
Understanding TC evolution in the lead-up to RI events is therefore a critical component of forecasting and damage mitigation efforts. In this work, we leverage powerful machine learning (ML) methods to provide \emph{scientific insight} into the evolution of TCs leading up to and during such events.

Multiple internal and external factors drive TC behavior \cite{kaplan2015rii}. This work focuses on a key internal factor: the spatio-temporal structure of deep convection---that is, vertical circulation of air---within the storm. Convective patterns within TCs serve as reliable indicators of intensity and drivers of intensity change; for example, identifying convective patterns is critical to the Dvorak Technique for TC intensity estimation \cite{Dvorak1975,olander2019advanced,ritchie2014satellite,hu2020short}. Since temperature generally decreases with height in the troposphere, the temperature of a cloud top is a proxy for its altitude. Thus taller clouds have colder tops, implying deeper (and thus stronger) convection. Stronger convection, particularly near the TC center, tends to be associated with stronger TCs. Infrared (IR) imagery from geostationary satellites such as the Geostationary Operational Environmental Satellites (GOES) provide high spatio-temporal resolution estimates of cloud-top temperature ($10.4\mu m$ wavelength) \cite{schmit2017closer}.

\begin{figure}[t!]
\floatbox[{\capbeside\thisfloatsetup{capbesideposition={left,top},capbesidewidth=.31\textwidth}}]{figure}[\FBwidth]
{\caption{\textbf{From Observation to Machine Learning to Science.} 
Meteorologists seek to connect observations and summaries of TC evolution to processes such as RI (blue path). We pose this scientific question as a hypothesis test (bottom right). By casting the hypothesis test as a prediction problem, we are able to leverage high-capacity AI techniques, here ANNs (top right). Such black-box models often struggle to extract scientifically meaningful insight (red path). Our framework facilitates insight on labeled sequences of functional data via local results in the original space of scientific functional features (black path).}\label{fig:flowchart}}
{\includegraphics[width=.62\textwidth]{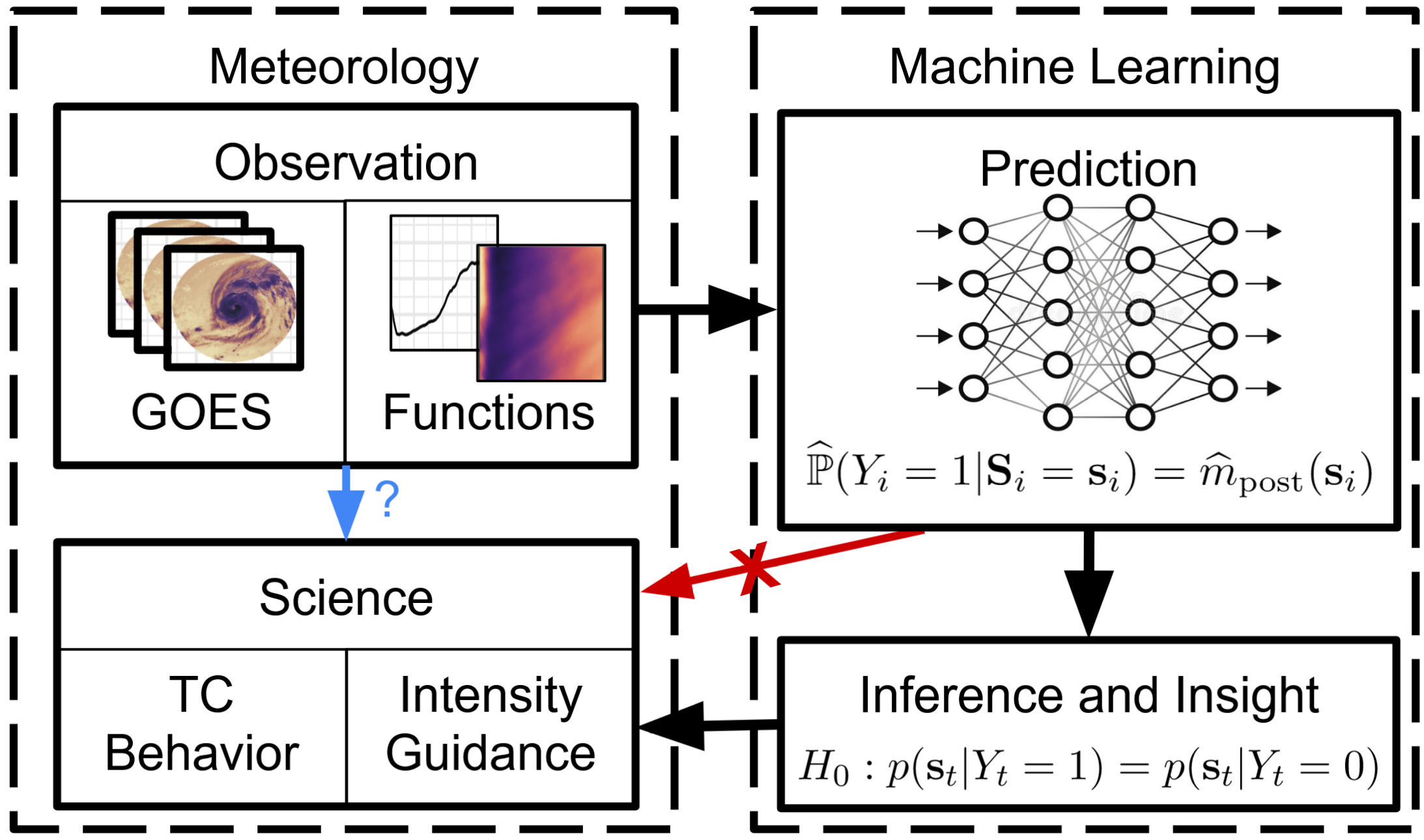}}
\end{figure}

\textbf{Scientific Goals.} TC intensity forecasts are ultimately issued by human experts who synthesize domain knowledge with the outputs of many forecasting tools to issue a single forecast. It is thus unproductive to provide another statistical model which only predicts TC intensity directly from IR imagery; such black-box models are difficult for forecasters to utilize. Our goal is instead to provide meteorologists with an enhanced understanding of how convective evolution relates to RI events ({Figure~\ref{fig:flowchart}}, blue arrow). Currently operational statistical forecast tools are generally limited to a handful of scalar summary statistics from each GOES image to aid forecasters in quickly digesting convective behavior \cite{DeMaria1999,demaria2009simplified,kaplan2015rii}. Conversely, we work with the rich suite of ORB functions, a dictionary of scientifically meaningful one-dimensional functional features \cite{mcneely2020unlocking}. In this paper, we illustrate our methods on one such ORB function---the radial profile---but our framework can generalize to jointly analyze multiple functional inputs. The radial profile, defined as $\overline{T}(r)=\frac{1}{2\pi}\int_0^{2\pi}T_b(r,\theta)d\theta$, captures the structure of cloud-top temperatures $T_b$ as a function of radius $r$ from the TC center and serves as an easily interpretable description of the depth and location of convection near the TC core \cite{mcneely2020unlocking,Sanabia2014}. A single profile describes the instantaneous convective structure ($\X_t$; Figure \ref{fig:data}, center) of a TC, whereas a 24-hour (48 observation) sequence of profiles provides a concise, human-readable summary of evolution over time; we call the latter sequence a \emph{structural trajectory} ($\S_t=\{\X_t,\X_{t-1},\dots,\X_{t-48}\}\in\mathcal{S}$; Figure \ref{fig:data}, right).

\textbf{Contribution.} We seek to answer the meteorological questions: ``Are the sequences $\S_t$ leading up to RI events different from those leading up to non-RI events?'' and, if so, ``How do RI events differ from non-RI events over $\mathcal{S}$?'' We pose this question as a hypothesis test comparing the distributions of $\S_t$ on $\mathcal{S}$ between classes ({Equation~\ref{eqn:original_hypothesis}}). By recasting the hypothesis as a prediction problem, we are able to leverage artificial neural networks (ANNs) to \textbf{assess the relationship between labeled sequence data} $\series{(\S_t,Y_t)}$ and TC behavior. Within this framework, shown in Figure \ref{fig:flowchart}, we not only detect the differing behavior of convection leading up to RI, but we are able to identify locations in $\mathcal{S}$ corresponding to RI, thus \textbf{bridging the gap from ML back to meteorological insight} ({Figures~\ref{fig:results},\ref{fig:case-study}}). To our knowledge, this is the only principled framework that can detect distributional differences in sequences of images, and in addition provide local diagnostics which can identify spatio-temporal patterns commonly found in the lead-up to RI events.

\section{Data} \label{sec:data}

\begin{figure}[b!]
\floatbox[{\capbeside\thisfloatsetup{capbesideposition={left,top},capbesidewidth=.3\textwidth}}]{figure}[\FBwidth]
{\caption{\textbf{Structural Trajectories as Image Data.} The raw data for each trajectory $\S_t$ is a sequence of cloud-top temperature images from GOES ($T_b$). We convert each of these into a radial profile ($\mathbf{X}_t$). The 24-hour sequence of radial profiles are combined into a \emph{structural trajectory} or image ($\S_t$). These images are the high-dimensional inputs to  $\widehat{m}_\text{post}(\s_t)$.}\label{fig:data}}
{\includegraphics[width=.64\textwidth]{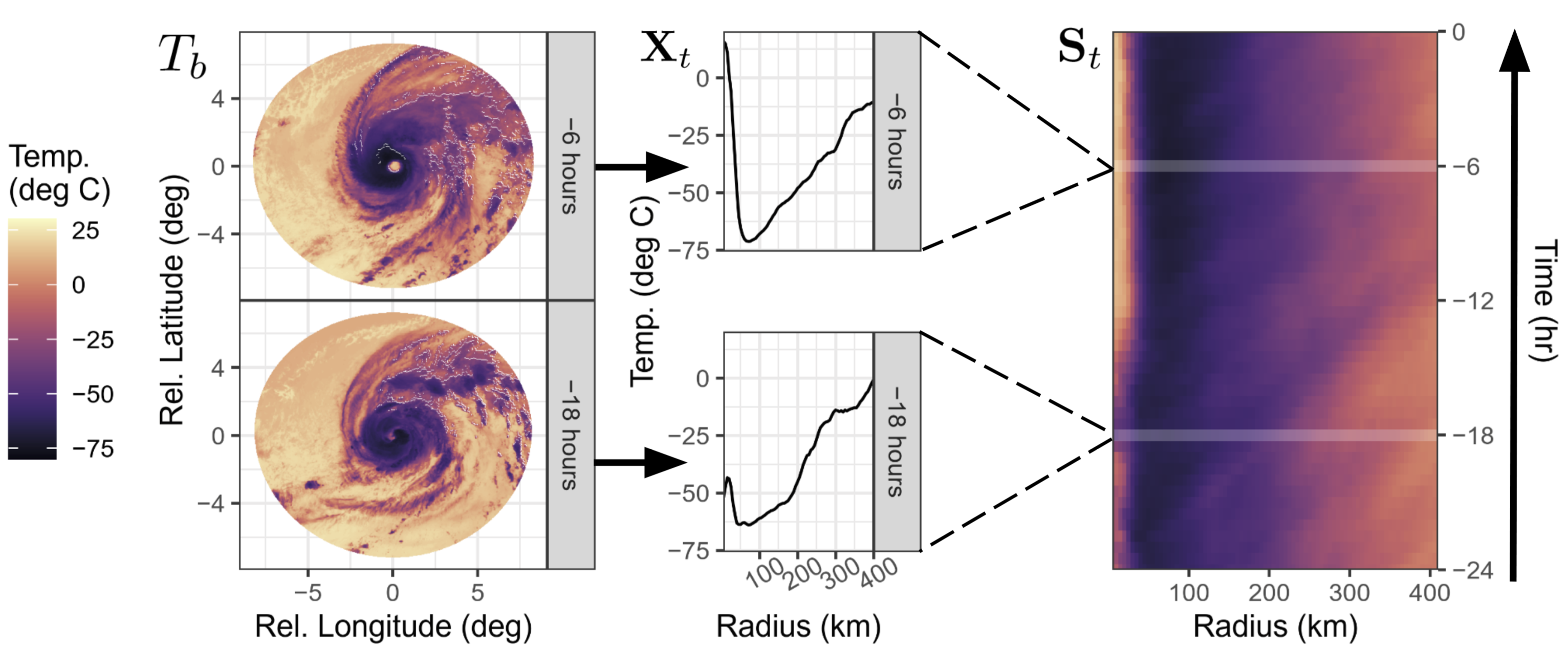}}
\end{figure}

\textbf{Inputs.} Raw GOES image data are drawn from NOAA's MERGIR database \cite{mergir}, which contains 30-minute$\times$4-km resolution IR imagery dating back to 2000 (Figure \ref{fig:data}, left). The NOAA HURDAT2 database provides TC best track data: the best estimate of TC intensity and location produced after each season using all available data, including data not available in real time \cite{Landsea2013}. HURDAT2 has a 6-hour resolution.

\textbf{RI Labels.} We attach labels $Y_t \in \{0, 1\}$ to structural trajectories $\S_t$ at each synoptic time (0000 UTC, 0600 UTC, 1200 UTC, and 1800 UTC) based on whether the TC was undergoing RI at time $t$ (where $Y = 1$ indicates RI) according to Algorithm \ref{alg:RI}. The non-synoptic times are labeled $Y_t=1$ if they fall between two consecutive synoptic $Y_t=1$ observations, and $Y_t=0$ otherwise.

\textbf{Sample Size.} We train our ANN on 33,873 labeled structural trajectories from 2000-2012 (further divided into 60\% train/40\% validation). We hold out 14,447 labeled trajectories from 2013-2019 as test data. When only labels $Y_t$ --- and not trajectories $\S_t$ --- are required ($\widehat{m}_\text{seq}$; Alg. \ref{alg:distributional-differences}(3)), we train on 31,203 synoptic best track entries from 1979-2012; we did not use data prior to 1979 due to reduced availability of satellite observations to support post-season analysis and thus the HURDAT2 best track database.

\section{Methods} \label{sec:methods}
\textbf{Testing Hypotheses.} Our goal is to  detect distributional differences in labeled sequence data $\series{(\S_t, Y_t)}$, where the ``labels'' $Y_t \in \{0,1\}$ are binary, and each covariate $\S_t \in \mathcal{S}$ is a sequence of functional quantities (our structural trajectory, Figure \ref{fig:data}). Formally, we test the hypothesis
\begin{align}
   H_0: p (\s_t | Y_t=1) &= p (\s_t | Y_t=0)\ \text{for all} \ \s_t \in \mathcal{S}   \ \text{versus}\label{eqn:original_hypothesis}\\\ H_1: p (\s_t | Y_t=1) &\neq p (\s_t | Y_t=0) \ \text{for some} \ \s_t \in \mathcal{S}.\notag
\end{align}
In our setting, there are two methodological challenges: (i) Equation \ref{eqn:original_hypothesis} is a high-dimensional two-sample test for dependent sequence data, and (ii) to glean scientific insight, an answer of whether we reject/fail-to-reject $H_0$ is not enough: if we reject $H_0$, then we need to identify the \emph{regions} in $\mathcal{S}$ where the two distributions differ.

\textbf{From Tests to Prediction.} For independent and identically distributed (IID) data $\{\S_i\}_{i = 1}^{n}$, {Equation~\ref{eqn:original_hypothesis}} can be re-cast as a classification problem via Bayes Theorem, resulting in an equivalent test of ${H_0: \P(Y_i = 1 | \s_i) = \P(Y_i = 1), \: \forall \, \s_i \in \mathcal{S}}$. 
This idea is at the core of the recent classification accuracy \cite{kim2021classification} and regression-based \cite{kim2019global} two-sample tests. The latter regression test goes beyond classification to provide \emph{local} results, indicating where in $\mathcal{S}$ the two distributions differ. The key quantity to estimate in this procedure is the \emph{class posterior} $m_\text{post}(\s_i) := {\P}(Y_i=1|\s_i)$. For this, one can leverage any probabilistic classification technique; hence converting a challenging inference and testing problem to a prediction problem. We reject $H_0$ if the test statistic $\lambda:=\sum_{t\in\mathcal{V}}(\widehat{m}_\text{post}(\s_i) -\widehat{m}_\text{prior})^2$  is large on a held-out evaluation set $\mathcal{V}$ (not used in training). Here $\widehat{m}_\text{prior}$ is the proportion of $Y_i=1$ cases in the train sample, and $\widehat{m}_\text{post}$ is the estimated class posterior.

\textbf{Accounting for Dependence.} However, existing high-dimensional tests do not extend to dependent sequence data. In particular, the regression test \cite{kim2019global} relies on a permutation test and the assumption of exchangeability --- an assumption that is generally violated for spatio-temporal processes. Here we propose a generalization of the regression test from IID to dependent data by replacing the permutation test with a {\em Monte Carlo (MC) test}. Our MC test estimates the label distribution $m_\text{seq}(y_{t-1},\dots,y_1) := \P(Y_t=1|Y_{t-1}=y_{t-1},\dots,Y_1=y_1)$ under $H_0$ in place of permutation. The MC test is asymptotically valid (that is, it controls Type I errors), as long as the estimator $\widehat{m}_\text{seq}$ converges to $m_\text{seq}$. See Algorithm \ref{alg:distributional-differences} for a full description of the testing procedure.

\textbf{Local Importance.} If we reject $H_0$, then the magnitude of the terms $\lambda_t:=\widehat{m}_\text{post}(\s_t)-\widehat{m}_\text{prior}$ in the test statistic $\lambda$ provide information as to which types of structural trajectories contributed the most to the rejection. Hence, we assign a local {\em importance score} $I(\s_t):=\text{sign}(\lambda_t)/\widehat{p}_t$ to each sequence $\s_t$, where $\widehat{p}_t$ is an approximate local $p$-value associated with $\lambda_t$, computed according to Algorithm \ref{alg:distributional-differences}.  Positive scores $I(\s_t)\gg0$ indicate $p(\s_t|Y_t=1)>p(\s_t|Y_t=0)$, and negative scores $I(\s_t)\ll0$ the reverse. In other words, our test and diagnostics can identify 24-hour trajectories which are more common to the lead-up to RI events.

\section{Results}\label{sec:results}
\textbf{Testing Hypotheses.} We execute Algorithm \ref{alg:distributional-differences}, estimating $m_\text{seq}$ with a Markov chain model with 48-hour memory and $m_\text{post}$ with a convolutional neural network (CNN). The CNN classifier achieved $\sim$75\% balanced accuracy on test data; note that classifier performance affects the power of the test, but not validity. We find that the distributions of RI vs non-RI trajectories are different at significance level $\alpha=0.01$.

\begin{figure}[h!]
    \centering
    \includegraphics[width=\textwidth]{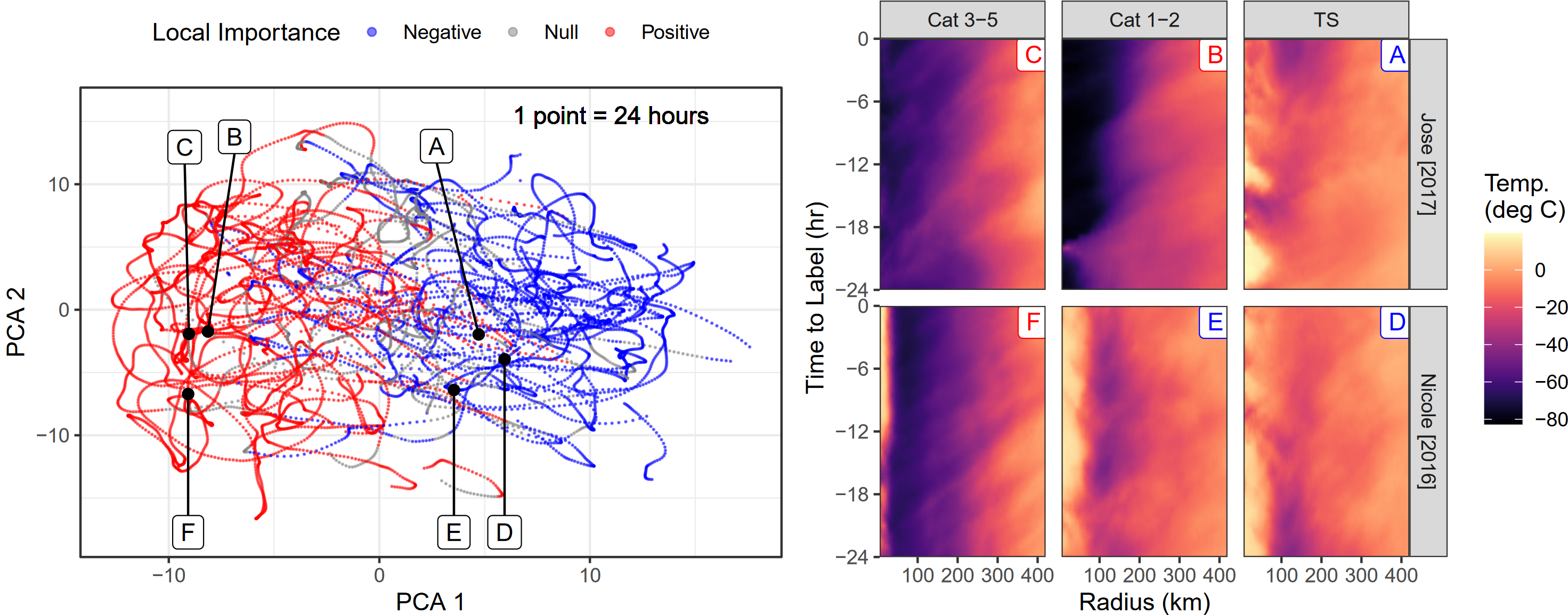}
    \caption{\textbf{Differences in Distribution.} (\emph{Left}) PCA map, where each point, colored by its importance score,  represents a 24h structural trajectory. In general, importance decreases from left to right, while current intensity increases from top to bottom. Note that PCA is only used for visualization purposes; the test is performed on the entire trajectory $\S_t$. (\emph{Right}) Panels of selected trajectories or points for Hurricane Jose [2017] and Hurricane Nicole [2016]. The inset letter indicates the location and local importance in the PCA map. Positive cases tend to have cold cloud tops near the core, growing in coverage and depth of convection with time.}
    \label{fig:results}
\end{figure}

\textbf{Local Importance.} Figure \ref{fig:results} (left) projects $\S_t \in \mathcal{S} \subset \R^{48 \times 120}$ onto their first two principal components. We use principal component analysis (PCA) for visualization only; $\widehat{m}_\text{post}$ is trained on the full structural trajectories. In Figure \ref{fig:results} (left), a single point represents an \emph{entire} 24-hour trajectory. Temporally adjacent trajectories are highly correlated in PCA space, as evidenced by the ``strands" in Figure \ref{fig:results}. We divide the importance scores $I(\s_t)$ for each trajectory into 3 categories: negative ($I(\s)<-40$; blue, RI less likely), null ($-40 < I(\s) <40$; gray), and positive ($I(\s)>40$; red, RI more likely). Positive importance scores tend to have low values of the first principal component score, corresponding to cold trajectories whose temperatures are decreasing.

\textbf{Scientific Insight.} Six sample trajectories from two example TCs (Hurricanes Nicole [2016] and Jose [2017]) are labeled in Figure \ref{fig:results} (left) and displayed in Figure \ref{fig:results} (right). The PCA map clearly show two patterns: (i) core convection is stronger for low values of PCA1 and (ii) eye-eye wall structures are present for low values of PCA2. 

We show the raw images $\S_i$ for each point in Figure \ref{fig:results} (right) along with Saffir-Simpson intensity categories. Out method is able to identify the structural patterns which lead up to RI: In all three examples with positive importance scores (B, C, F), cloud-top temperatures are cold and getting colder. Trajectory F also shows the formation of an eye, during which time the TC intensified by 40 kt. The examples with negative importance scores (A, D, E) have warmer cloud-top temperatures with little indication of cooling in the core.  \textbf{This analysis corroborates other studies that indicate strong, deepening convection is crucial for RI, particularly near the TC core where energy is most likely to result in strengthening surface winds.}\cite{rogers2013airborne}

\begin{figure}[h!]
    \centering
    \includegraphics[width=\textwidth]{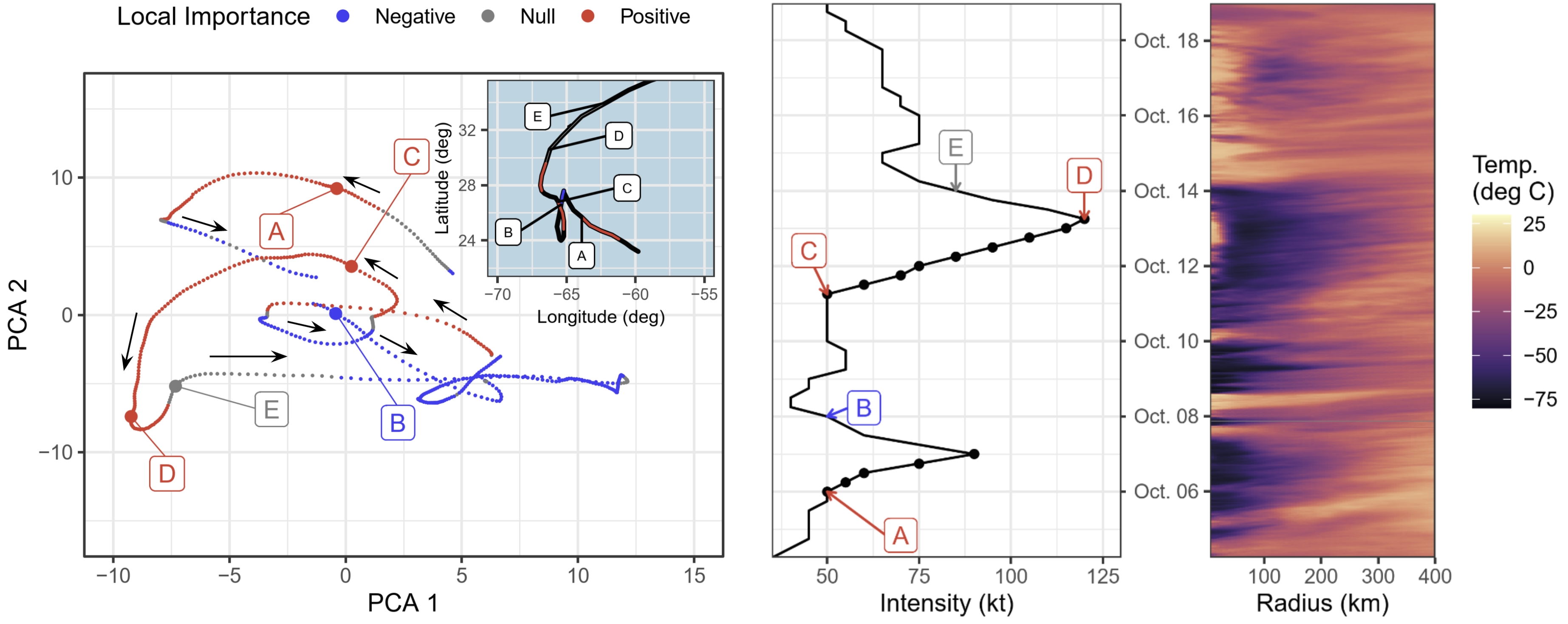}
    \caption{\textbf{Case Study: Hurricane Nicole [2016].} (\emph{Left}) Full set of trajectories $\series{\S_t}$ for Hurricane Nicole [2016], with select trajectories labeled; again, each point represents a 24-hour trajectory. The physical track of the storm across the North Atlantic is shown inset, with the same trajectories labeled (here represented as segments of the track). (\emph{Center}) Storm intensity over time, where solid points mark RI events. (\emph{Right}) Full trajectory of Hurricane Nicole over $~$2 weeks. Note that the importance scores capture rapidly intensifying periods (A, C, D), collapsing convection (B), and the decay of the TC eye (E).}
    \label{fig:case-study}
\end{figure}

\textbf{Case Study: Hurricane Nicole [2016].} We examine the full series $\series{(\S_t,Y_t)}$ for Hurricane Nicole [2016] in Figure \ref{fig:case-study}. When $I(\s_t)\gg0$, we indeed find that TC intensity is increasing (center; points A, C, D). More interesting still, the trajectories \emph{prior} to points A and C --- the \emph{onset} of RI --- already begin to indicate that RI is likely; there are obviously markers of RI in the radial profiles which may be of use to forecasters. Future work will identify archetypal trajectories of this sort: $I(\s)\gg0$ prior to intensification.

\section{Conclusion} 
Our method successfully detects distributional differences in 24-hour ``structural trajectories'' (derived from GOES-IR images) leading up to periods of rapid intensification. A preliminary partitioning of trajectories by our proposed ``local importance score'' shows several known archetypal structural trajectories, laying the groundwork for forecasting tools. Further extending this framework to trajectories of other high-dimensional data promises to provide the insight necessary to understand and predict these increasingly deadly events.

\section{Acknowledgments} 
This work is supported in part by NSF DMS-2053804, and the NSF AI Institute: Physics of the Future, NSF PHY-2020295. K. M. Wood was supported by the Mississippi State University Office of Research and Economic Development. R. Izbicki was supported by FAPESP (grant 2019/11321-9) and CNPq (grant 306943/2017-4).

\bibliographystyle{plain}
\bibliography{main}

\appendix
\section{Algorithms}

\begin{algorithm}[h!] 
	\caption{Test for distributional differences in labeled sequence data.}
	\label{alg:distributional-differences}
	{\small
		\begin{algorithmic}
			\REQUIRE type of test (MC=TRUE for MC test; MC=FALSE for permutation test); train data $\{Y_t, \S_t\}_{t \in \mathcal{T}_1}$ and regression method for estimating $m_\text{post}(\s):= \mathbb{P} (Y_t=1 | \mathbf{S}_t=\mathbf{s})$; for MC test, additional train data $\{Y_t\}_{t \in \mathcal{T}_2}$ and regression method for estimating $m_\text{seq}(Y_{t-1},\dots,Y_{t-k}):=\P(Y_t=1|Y_{t-1},\dots,Y_{t-k})$; number of repetitions $B$; test points $\{\s_t\}_{t \in \mathcal{V}}$.  \vspace{0.25em}\\
			\ENSURE p-value for testing $H_0: p (\s_t | Y_t=1) = p (\s_t | Y_t=0),\ \text{for all} \ t$ and $\s_t$, local importance scores $\{I(\s_t)\}_{t\in\mathcal{V}}$
			\vspace{0.7em}\\
			
			\STATE \texttt{// Estimate underlying probability distributions}\vspace{0.25em}

			\STATE (1) Estimate $m_\text{prior} := \P(Y_t = 1)$ with 
			class proportion $\widehat{m}_\text{prior}=\frac{1}{\lvert\mathcal{T}_1\rvert} \sum_{t\in\mathcal{T}_1}Y_t$.\vspace{0.25em}
			
			\STATE (2) Regress $Y_t$ on $\S_t$ using $\mathcal{T}_1$ to compute $\widehat{m}_\text{post}$. \vspace{0.25em}

			\STATE (3) \textbf{if} {MC} \textbf{then} {Regress $Y_t$ on $Y_{t-1},\dots,Y_{t-k}$ using $\mathcal{T}_2$ to compute $\widehat{m}_\text{seq}$.}
		
	  	\STATE \texttt{// Compute test statistic and estimate its null distribution}	\vspace{0.25em}
			\STATE (4) Compute test statistic $\mathbf{\lambda}=\sum_{t\in\mathcal{V}}\lambda_t^2,\text{ where }\lambda_t=\widehat{m}_\text{post}(\s_t)-\widehat{m}_\text{prior}$.\vspace{0.25em}	
			\STATE (5) \For{$b\in \{1,2,\dots,B\}$}{
			\STATE - Draw new train labels $\{\widetilde{Y}_t\}_{t\in\mathcal{T}_1}$ under $H_0$:
		   \IF{MC} {\STATE - Initialize $\widetilde{Y}_{1},\dots,\widetilde{Y}_{k}$ to random sequence.
		            \STATE - Draw sequence of length 100$\times k$ from $\widetilde{Y}_t\sim\text{Binom}(\widehat{m}_\text{seq}(\widetilde{Y}_{t-k},\dots,\widetilde{Y}_{t-1}))$ for burn-in.
		            \STATE - Draw new labels $\widetilde{Y}_t\sim\text{Binom}(\widehat{m}_\text{seq}(\widetilde{Y}_{t-k},\dots,\widetilde{Y}_{t-1}))$, for $t \in \mathcal{T}_1 \setminus \mathcal{T}_0$.}
		            \ELSE {\STATE - Permute original labels $\{Y_t\}_{t \in \mathcal{T}_1}$.}
		   \ENDIF

			\STATE - Regress $\widetilde{Y}_t$ on $\S_t$ using $\mathcal{T}_1$ to compute $\widehat{m}_\text{post}^{(b)}$.\vspace{0.25em}
			
			\STATE - Recompute test statistic $\widetilde{\lambda}^{(b)}=\sum_{t \in \mathcal{V}}\widetilde{\lambda}_t^{(b)2}\text{ where }\widetilde{\lambda}_t^{(b)}= \widehat{m}^{(b)}_\text{post}(\s_t)-\widehat{m}_\text{prior}$.
			}
			
			\STATE \texttt{// Compute approximate p-values}	\vspace{0.25em}
			
			\STATE (6) Compute $p$-values according to
	            {$$\widehat{p}_\text{global} = \frac{1}{B+1} \left( 1 + \sum_{b=1}^B I\left(\widetilde{\lambda}^{(b)} > \lambda\right) \right)\text{ and }\widehat{p}_t=\frac{1}{B+1} \left( 1 + \sum_{b=1}^B I\left(|\widetilde{\lambda}_t^{(b)}| > |\lambda_t|\right) \right).$$}
	            
	        \STATE (7) Define local importance scores by $$I(\s_t)=\frac{\text{sign}{(\lambda_t)}}{\widehat{p}_t}$$
	            
	        \STATE \textbf{return} $\widehat{p}_\text{global},\ \{I(\s_t)\}_{t\in\mathcal{V}}$
	        
		\end{algorithmic}
	}
\end{algorithm}

\begin{algorithm}[!t]
	\SetAlgoLined
	\SetKw{req}{Require:}
	\SetKw{output}{Output:} 
	\req Sequence of contiguous intensity (i.e. maximum wind speed) observations $\{W_i\}_{t=1}^T$ for a storm and an intensity change threshold $c$. \\
	\textbf{Initialize:} $\mathbf{Z}=\mathbf{0}_{T,T}$, $Y=\mathbf{0}_T$\\
	\For{$t$ in $1:T-1$}{
	$\Delta_{t,1}=W_{t+1}-W_t$; \texttt{// lead-1 intensity change}\\ 
	  \If{$t\le T-4$}{
	  	$\Delta_{t,4}=\max(W_t,\dots,W_{t+4})-W_t$; \texttt{// lead-4 intensity change}
  	  }
	}
	$\mathcal{A}=\{t:\Delta_{t,4}\ge c\}$; \texttt{// 24-hour windows containing RI}\\
	$\mathcal{B}=\{t:\Delta_{t,1}>0\}$; \texttt{// 6-hour windows containing intensification}\\
	
	\For{$t$ in $1:T-4$}{
	  \If(\texttt{// label 24-hour RI window}){$t\in\mathcal{A}$}{
		$Z_{t,t},\dots,Z_{t,t+4}=1$;\\
		\For{$h$ in $4:1$}{
		  \uIf(\texttt{// trim non-intensification from end of event}){$t+h-1\notin\mathcal{B}$}{
			$Z_{t,t+h}=0$;\\
		  }\Else{
	        break;
          }
		}
	    \For{$h$ in $0:3$}{
	      \uIf(\texttt{// trim non-intensification from start of event}){$t+h\notin\mathcal{B}$}{
	        $Z_{t,t+h}=0$;\\
	      }\Else{
	    	break;
	      }
	    }
	  }
	}
	\For{$t$ in $1:T$}{
	  $Y_t=\max(Z_{t,1},\dots,Z_{t,T})$; \texttt{// points only need to be valid for one start}
    }
	\output $\{Y_i\}_{t=1}^T$\\
	\textbf{Note:} {\em The above algorithm identifies rapid intensification. To identify rapid weakening instead, reverse the input sequence $\{W_i\}_{t=1}^T$ at initialization, then reverse the output sequence $\{Y_i\}_{t=1}^T$ at output.}
	\caption{Identifying observations that fall within rapid intensification events.}
	\label{alg:RI}
\end{algorithm}

\end{document}